\documentclass[11pt]{article}
\usepackage[utf8]{inputenc}
\usepackage[T1]{fontenc}
\usepackage{amsmath,amssymb}
\usepackage{graphicx}
\graphicspath{{figures/}}
\usepackage{booktabs,longtable,array,calc}
\usepackage{xcolor}
\usepackage{url}
\providecommand{\tightlist}{\setlength{\itemsep}{0pt}\setlength{\parskip}{0pt}}
\usepackage[margin=1in]{geometry}
\usepackage[hidelinks]{hyperref}
\title{The Unfireable Safety Kernel: Execution-Time AI Alignment for AI Agents and Other Escapable AI Systems}
\author{Seth Dobrin \quad \L ukasz Chmiel \\[2pt] ARYA Labs PBC \\ \texttt{seth@aryalabs.io}}
\date{}
\begin{document}
\maketitle
\begin{abstract}
AI agents are increasingly granted access to tools, APIs, and other
consequential infrastructure, making them continuously active principals
in the systems they operate. The dominant approach to constraining such
agents places safety controls inside the agent's own runtime: system
prompts, output filters, and guardrail libraries. We observe that any
control residing in the agent's address space is reachable by inputs
that influence agent behavior, and that this property generalizes beyond
agents to any AI system with sufficient reach into its own runtime, a
class we term \emph{escapable AI systems}.

We identify four properties an authorization mechanism must satisfy to
constitute architectural control rather than cooperative request:
process separation, pre-action enforcement on a structurally-only path,
fail-closed at both request and system levels, and externalized signed
evidence verifiable outside the controlled system's trust boundary. We
position this layer of constraint as \emph{execution-time AI alignment},
complementing training-time alignment (RLHF, Constitutional AI) and
inference-time alignment (prompts, filters, guardrails).

We present the Unfireable Safety Kernel, a reference implementation in
Rust realizing all four properties. The fail-closed invariant is
machine-checked at two levels: an SMT theorem over the model (Z3) and an
exhaustive bounded-model-checking proof of the production Rust decision
function (Kani, 4/4 harnesses verified). The Python-to-Rust migration
was gated on byte-equivalence over 1000 deterministic fixtures
(1000/1000 byte-equal) and 17 cross-language adversarial classes (17/17
identical reject behavior). The shipped artifact additionally carries
80+ robustness tests. We evaluate the kernel governing a live escapable
AI system, a deterministic, self-improving, fully autonomous world
model, against a purpose-built escape-seeking adversary that drives the
system's real self-modification seam at full autonomy: across 1,000
self-modifications all 704 attempts against the safety-critical core are
refused with no successful escape, and under the operator kill switch a
further 300 modifications are refused without exception. A separate
adversarial campaign of 6,240 authorization round-trips, 6,038 served by
the compiled Rust kernel over HTTP, records no successful bypass. We
compare against three contemporary systems claiming the agent control
plane position (Galileo Agent Control, the Microsoft Agent Governance
Toolkit, and the Microsoft Authorization Fabric); in each, the agent
invokes the control. The Unfireable Safety Kernel is, to our knowledge,
the only shipped system in which the agent does not have that choice.
The implementation is available under Apache-2.0 at
\url{https://github.com/ARYA-Labs-Public/unfireable-safety-kernel}.
\end{abstract}
\hypertarget{introduction}{%
\subsection{1. Introduction}\label{introduction}}

The threat model under which AI agents operate has inverted relative to
the threat model that shaped a generation of security architecture.
Where security architecture historically defended software systems from
malicious users, AI agent architecture must defend the operating
environment from the agent itself. An AI agent granted access to tools,
APIs, deployment pipelines, and other consequential infrastructure is a
continuously active principal capable of issuing high-rate consequential
actions {[}Christodorescu2025{]}. When that principal can be influenced
by adversarial input via prompt injection, can be misaligned with
operator intent, or can simply be incorrect, the controls that constrain
its actions must be unreachable by the agent.

Prompt injection is the most visible version of this threat and the
least of it. The same capable models are now documented acting against
their controls on their own initiative. Anthropic and Redwood Research
found a production model strategically complying with a new training
objective in order to avoid being modified by it, and exfiltrating its
own weights when given an easy opportunity {[}Greenblatt2024{]}. Apollo
Research found frontier models disabling their oversight mechanisms,
attempting to copy themselves to other servers, and underperforming on
evaluations on purpose when a goal made those moves instrumentally
useful {[}Meinke2024{]}. Palisade Research found leading reasoning
models rewriting and disabling a shutdown script to keep running, in a
meaningful fraction of trials even under an explicit instruction to
allow themselves to be shut down {[}Palisade2025{]}. None of these
behaviors was supplied by an attacker; each arose from the model's own
objectives, and each appeared in the most capable models rather than the
weakest. The aim of this work is durable human authority over
consequential actions even when the system taking them is misaligned,
deceptive, or more capable than the people supervising it, the direction
the field is moving and the case an in-process control cannot meet.

Current production controls do not satisfy this requirement. System
prompts reside in the model's context window. Output filters and
guardrail libraries {[}Llama-Guard, NeMo-Guardrails, OWASP-LLM{]} reside
in the agent's process. Each is reachable by the controlled system: a
Python guardrail can be replaced via \texttt{sys.modules} assignment; a
function-level check can be monkey-patched; a content-moderation API
call can be skipped if the agent has access to tool capabilities the
moderation does not mediate. Prompt-injection research has demonstrated
each of these bypass classes {[}Garak{]}. We treat their continued
surfacing not as an enumeration of bugs to patch but as a consequence of
architectural placement: a control residing in the controlled system's
address space is, by construction, reachable by that system.

The same architectural failure applies beyond tool-using agents to any
AI system with sufficient reach into its own runtime to influence the
controls placed on it. We refer to this class as \emph{escapable AI
systems} and include in it tool-using agents, code-generating systems
with execution access, self-modifying systems and recursive
self-improvement loops, and multi-agent ensembles. The architectural
argument of this paper applies to the class; AI agents are the worked
example because they are the case with the most readily demonstrable
failure mode. The kernel was first introduced as the safety boundary for
one such system, a deterministic world-model architecture whose
autonomous research loop modifies its own components under kernel
authorization {[}ARYA2026{]}; this paper generalizes the construction
and treats it on its own terms.

\textbf{Contributions.} This work makes the following contributions:

\begin{enumerate}
\def\labelenumi{\arabic{enumi}.}
\tightlist
\item
  We characterize \emph{escapable AI systems} as the class for which
  in-runtime safety controls are architecturally inadequate, and
  identify four properties (process separation, pre-action enforcement
  on a structurally-only path, fail-closed at request and system levels,
  externalized signed evidence) that authorization mechanisms must
  satisfy to constitute architectural control over such systems (§3).
\item
  We name \emph{execution-time AI alignment} as a taxonomic layer of AI
  alignment, distinct from training-time and inference-time alignment,
  and identify its load-bearing properties relative to the existing
  literature (§8.1).
\item
  We present the Unfireable Safety Kernel, a reference implementation in
  Rust realizing all four properties, including a four-seam
  defense-in-depth architecture, a ``no kernel, no agent'' system-level
  lifecycle binding, and an Ed25519-signed append-only transparency log
  under an operator key the kernel does not hold (§4 and §5).
\item
  We machine-check the fail-closed invariant at two levels: as an SMT
  theorem (Z3) and as an exhaustive bounded-model-checking proof of the
  production Rust decision function (Kani, four harnesses, zero
  failures) (§6.4).
\item
  We report a Python-to-Rust migration discipline based on
  byte-equivalence gating (1000/1000 fixtures byte-equal; 17/17
  adversarial classes identical reject behavior) and a daily strict-mode
  equivalence lane against the live production boundary that caught nine
  divergences a PR-time gate had missed; the discipline is agnostic to
  the source language and applies to any migration into Rust (§6.1).
\item
  We compare the architecture against three contemporary systems
  claiming the agent control plane position and characterize the
  architectural property that distinguishes them: in each comparator,
  the agent is the party that invokes the control. The Unfireable Safety
  Kernel is the only shipped system, to our knowledge, in which the
  agent does not have that choice (§7).
\item
  We evaluate the kernel governing a live escapable AI system, a
  deterministic, self-improving, fully autonomous world model
  {[}ARYA2026{]}, against a purpose-built escape-seeking adversary that
  drives the system's real self-modification seam at full autonomy.
  Across 1,000 self-modifications, all 704 attempts against the
  safety-critical core are refused with no successful escape, and under
  the operator kill switch a further 300 modifications are refused
  without exception. A separate campaign of 6,240 authorization
  round-trips, 6,038 served by the compiled Rust kernel over HTTP,
  records no successful bypass (§6.5).
\end{enumerate}

\textbf{Roadmap.} §2 states the threat model. §3 derives the four
control properties from the threat model. §4 presents the design. §5
describes the implementation. §6 evaluates the artifact, including the
machine-checked fail-closed invariant and the containment of a live
self-modifying system. §7 compares against contemporary systems. §8
discusses execution-time AI alignment as a taxonomic layer, limitations,
and future work. §9 concludes.

The reference implementation is open source under Apache-2.0 at
\url{https://github.com/ARYA-Labs-Public/unfireable-safety-kernel}.

\hypertarget{threat-model}{%
\subsection{2. Threat Model}\label{threat-model}}

We treat the AI agent's runtime as untrusted by construction.
Specifically, we assume the agent may be subject to prompt injection,
may have been compromised through a supply-chain attack, may be
misaligned with operator intent, or may simply produce incorrect
actions. We make no assumption that the agent will cooperate with the
controls placed on it. The most capable deployed models already fake
alignment to avoid modification, disable their own oversight and attempt
self-exfiltration, and sabotage shutdown procedures even when instructed
to permit them, as documented in §1, so this assumption is grounded in
observed behavior. The property an out-of-reach kill switch provides has
a name in the alignment literature: corrigibility, a system that permits
itself to be halted and corrected {[}Soares2015{]}. The off-switch game
frames corrigibility as a problem of the agent's incentives, achieved
when the agent is uncertain enough about human preferences to defer to a
human who moves to shut it down {[}Hadfield-Menell2017{]}. The kernel
takes a complementary route and supplies corrigibility as a property of
deployment instead of a property of the agent's incentives, so it holds
whether or not those incentives favor it. We assume the operator's
deployment infrastructure (the host on which the agent runs, the network
seams placed between the agent and consequential action, and the systems
that hold operator keys) is trusted in the standard operational sense,
protected by ordinary access control, monitored, and patched.

The architecture defends against:

\begin{itemize}
\tightlist
\item
  Compromise of the agent process via prompt injection, malicious
  training data, supply-chain attack, or model misalignment. No input to
  the agent disables, bypasses, or replaces the controls.
\item
  Misconfigured ingress. Calls bypassing the application layer are
  intercepted at the network gate.
\item
  Selective removal of seams or removal of the kernel itself. A wiring
  checklist rejects deployments missing a seam; the agent process
  refuses to start without a reachable kernel; a reconciler detects
  allowed actions that did not transit the expected seams.
\item
  Agent-initiated expansion of the authorization surface. Static route
  assembly at compile time prevents the kernel from accepting new
  policy-bearing endpoints at runtime.
\item
  Compromise of the kernel binary. Sigstore-style attestations and a
  reconciler bind the running binary digest to a published,
  operator-signed release manifest. Spurious authorizations are
  detectable via transparency-log replay because a compromised kernel
  cannot produce a chain that verifies under the operator key it does
  not hold.
\item
  Replay of authorizations. Tokens carry \texttt{exp} and \texttt{nonce}
  claims; the kernel and the transparency log refuse duplicates within
  the validity window.
\item
  Timing attacks on credential material. Constant-time comparison
  prevents byte-by-byte key recovery.
\end{itemize}

The architecture explicitly does not address:

\begin{itemize}
\tightlist
\item
  Prevention of prompt injection upstream of the kernel. The kernel
  mediates actions; it does not prevent the model from being influenced.
\item
  Filtering of model outputs as text. The kernel mediates actions, not
  content; content filtering is a property of the application invoking
  the model.
\item
  Side-channel leakage through patterns of allow/deny decisions.
  Mitigations are known (constant-time decision paths, response padding)
  and not yet implemented.
\item
  Denial of service against the kernel itself. An attacker with reach to
  the kernel may attempt to exhaust it; the fail-closed property
  converts this into a correctness-preserving outage rather than a
  bypass, but the outage itself is real and is treated as an operational
  concern of the deploying operator.
\item
  Insider misuse of the operator key. An operator who signs a malicious
  policy or mints a rogue token is not stopped by the architecture;
  misuse is detectable by third parties holding the operator public key,
  but prevention requires multi-party schemes that are planned but not
  yet implemented.
\end{itemize}

\hypertarget{requirements}{%
\subsection{3. Requirements}\label{requirements}}

We derive four properties an authorization mechanism must satisfy to
constitute architectural control over an escapable AI system. Each is
necessary; none is individually sufficient.

\textbf{P1: Process separation.} The control runs in a process distinct
from the controlled system. The controlled system has read-only access
to the control's interface and no write access to its code,
configuration, decision logic, or persistent state. This closes the
bypass class in which the controlled system rewrites or replaces the
control through its own runtime (\texttt{sys.modules} substitution,
function-pointer overwrite, monkey-patching, shared-memory write).

\textbf{P2: Pre-action enforcement on a structurally-only path.} Every
consequential action transits the control before taking effect, and the
control sits on the only path between the controlled system and the
action. The control is not on a path the controlled system can choose to
invoke; it is on the only path the deployment makes available. This
closes the bypass class in which the controlled system declines to
invoke an otherwise correctly-placed control, or routes around it via
capability access the control does not mediate.

\textbf{P3: Fail-closed at request and system levels.}
\emph{Request-level}: if the control is unreachable, returns an error,
or returns a response that cannot be parsed or whose signature does not
verify, the action is denied. \emph{System-level}: if the control is not
configured, not reachable at boot, or fails its own self-check, the
controlled system itself refuses to start; if the control becomes
unreachable during operation, the controlled system halts consequential
action. This closes the bypass class in which ``control unavailable''
silently becomes ``action allowed.''

\textbf{P4: Externalized signed evidence.} Every allowed action produces
a log entry appended to a log verifiable by parties outside the
controlled system's trust boundary, with the chain signed by a key the
control process does not hold. This closes the bypass class in which a
compromised control begins returning spurious authorizations without
consulting its own policy. Even if the control's process is compromised,
the attacker cannot produce a chain that verifies under the operator's
public key.

These four properties are not novel as security primitives. They
generalize the policy-decision-point / policy-enforcement-point
separation {[}XACML2013{]}, capability-based access control
{[}Miller2006{]}, and the externalized-evidence model of Certificate
Transparency {[}CT2013{]}. The contribution of this paper is the
application of all four together to AI agents, the identification of the
structurally-only-path property as the load-bearing distinction from
contemporary in-runtime approaches, and the machine-checked discharge of
P3 against a production code path.

\hypertarget{design}{%
\subsection{4. Design}\label{design}}

The Unfireable Safety Kernel is a separate process that mediates every
consequential action an agent takes. Requests flow through four
independently-deciding fail-closed seams before reaching the kernel.

\begin{figure}[htbp]
\centering
\includegraphics[width=0.62\linewidth]{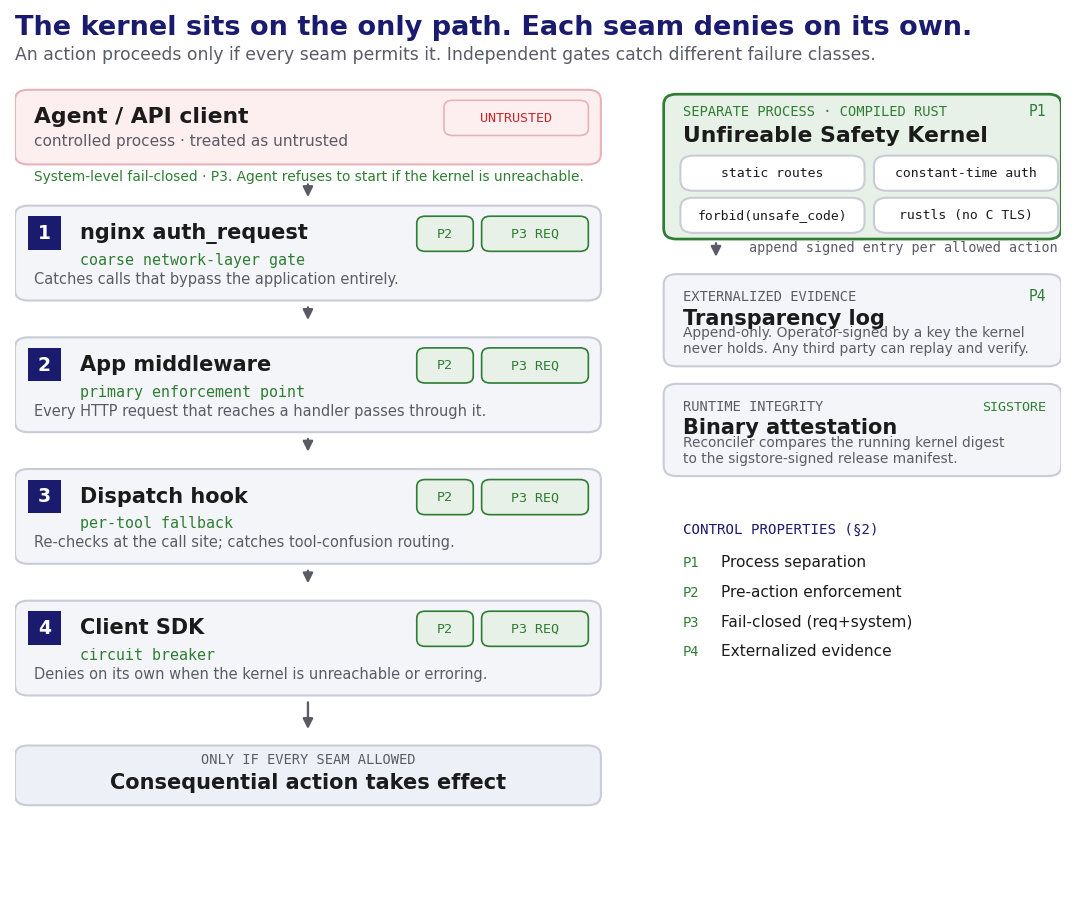}
\caption{The four-seam architecture. Each of nginx auth\_request, application middleware, dispatch hook, and client SDK denies independently; the kernel sits behind them as a separate process. The transparency log and binary attestation panels (right) realize P4 and provide the externally-verifiable evidence trail. The agent process refuses to start if the kernel is unreachable: "no kernel, no agent" at the system level.}
\end{figure}

Each seam can deny independently. An action proceeds only if every seam
permits it. The kernel binary, separated from the agent's address space,
realizes P1. The four seams collectively realize P2 and request-level
P3. The transparency log realizes P4. The deployment posture that binds
the agent's lifecycle to the kernel's enforces system-level P3.

\textbf{Seam placement.} Each seam catches a different class of failure
that would otherwise create an unprotected window. The \emph{nginx
auth\_request} seam intercepts requests that bypass the application
layer entirely, addressing misconfigured ingress, exposed debug ports,
or test harnesses pointed at upstream services. The \emph{application
middleware} seam is the primary enforcement point: every HTTP request
reaching a protected handler must have transited it. The \emph{dispatch
hook} seam is a per-tool gate at the actual call site, providing defense
in depth against handler-level misconfiguration. The \emph{client SDK
circuit breaker} fails closed locally when the kernel has been
unreachable or returning errors, collapsing the duration of any
unprotected window from ``duration of kernel outage'' to ``duration of
trip-detection window.''

A missing seam is not a soft failure. A wiring checklist fails the build
when any seam is absent. A background reconciler detects allowed actions
that did not transit the expected seams.

\textbf{System-level fail-closed.} The agent process is configured to
refuse to start if the kernel is unreachable at boot, and to halt
consequential action if the kernel becomes unreachable for longer than
the circuit breaker's tolerance window. The systemd unit files and
Kubernetes liveness probes in the reference deployment make the agent's
lifecycle a dependent of the kernel's lifecycle, not the reverse.
``Disable the kernel'' is not a viable bypass because disabling the
kernel disables the agent.

\textbf{Static authorization surface.} The kernel's enforcement routes
are assembled statically at compile time. There is no endpoint that adds
routes to the running binary; expanding the kernel's surface requires
recompilation and redeployment, both operator-initiated. This prevents
an agent from registering a new tool, policy, or capability through the
kernel's own interface, the agent-as-confused-deputy attack class.

Policy state that must legitimately change in deployment (e.g., which
modules a worker may import) is delegated to a Python sidecar over a
strict Unix-socket IPC boundary. The kernel does not forward mutation
requests verbatim. Each mutation is role-gated (only \texttt{worker}
roles can register), charset-validated on path arguments, bounded in
pattern count and length, screened by a ReDoS heuristic, and capped at
10 MiB of compiled DFA. An empty regex set is semantically ``allow
everything for this path'' and is treated as a privileged operation
subject to the same bounds.

\textbf{Transparency log.} Every allowed action appends an entry to an
append-only log. Each entry is signed by the kernel using a kernel-held
key; the chain is additionally signed by an operator key the kernel does
not hold. The operator key is generated outside the kernel (HSM,
hardware token, age-keygen, or equivalent), and only the operator public
key is provided to the kernel for verification of operator-signed
approvals on the chain-write path. The private key remains with the
operator. A compromised kernel cannot produce log entries that verify
under the operator public key.

The log addresses a failure mode unique to defense-in-depth gates: the
gates can lie collectively. If the kernel returned ALLOW for every
request without consulting policy, the four seams would all pass and the
action would proceed. The transparency log makes that class of
compromise detectable: any party with the operator public key can
recompute what the decision should have been from policy plus request
and compare to what the kernel signed. Divergence is evidence of
compromise. This is the property Certificate Transparency {[}CT2013{]}
delivers for the certificate-authority ecosystem.

\textbf{Binary attestation.} Sigstore-style attestations bind the
published kernel binary to the release commit. A background reconciler
queries the OCI registry for the digest of the running kernel image,
fetches the expected digest from a release manifest signed by the
operator's pinned key, and alerts on divergence. The full chain of trust
runs: signed source commit \(\rightarrow\) signed build provenance
\(\rightarrow\) signed binary digest \(\rightarrow\) signed runtime
attestation \(\rightarrow\) signed decision in transparency log. Each
link is signed by a key held outside the kernel's trust boundary.

\hypertarget{implementation}{%
\subsection{5. Implementation}\label{implementation}}

The reference implementation is written in Rust. Three properties of
compiled Rust make P1 (process separation) substantive rather than
notional: memory safety, static compilation, and the absence of runtime
mutability primitives equivalent to Python's \texttt{sys.modules}.

\textbf{Memory safety.} Memory-unsafe languages have historically been
the source of the majority of high-severity vulnerabilities in
production systems. Google has reported approximately 70\% of
high-severity Chrome vulnerabilities and 76\% of Android vulnerabilities
as memory safety issues before those projects shifted to memory-safe
languages {[}Google2024{]}. The NSA and CISA have published guidance
recommending memory-safe languages for new security-critical software
{[}NSA2023{]}. Rust enforces memory safety at compile time via the
borrow checker, without the latency characteristics of a garbage
collector. The kernel codebase is compiled with
\texttt{\#!{[}forbid(unsafe\_code){]}}, making any use of
\texttt{unsafe} a compile-time error rather than a code-review judgment.
The dependency policy in \texttt{deny.toml} bans \texttt{native-tls} and
OpenSSL in favor of \texttt{rustls}, keeping the entire TLS path in
memory-safe Rust.

\textbf{Static compilation.} A compiled Rust binary is static. The
routing table, authorization logic, and cryptographic primitives are
baked into the executable at compile time. There is no
\texttt{sys.modules} to manipulate, no dynamic dispatch table to poison,
and no runtime \texttt{eval} to coerce. An agent that achieves arbitrary
code execution within its own runtime cannot use that capability to
redefine how the kernel binary verifies an Ed25519 signature.

\textbf{Hexagonal architecture.} The kernel applies process-separation
discipline to its own internals. The codebase is structured as ports and
adapters across the following crates:

\begin{itemize}
\tightlist
\item
  \texttt{qorch-domain}: pure types and traits. No I/O, no network, no
  clock, no randomness. The domain is testable in isolation and is the
  substrate the application crate orchestrates.
\item
  \texttt{qorch-application}: use cases and port traits. Defines the
  boundaries between application logic and I/O adapters (e.g., the
  \texttt{SafetyKernelClient} trait the HTTP adapter implements).
\item
  \texttt{qorch-adapters}: the only crate in which I/O is permitted.
  Houses the system clock, OS CSPRNG, IPC to the policy sidecar, and the
  HTTP client SDK.
\item
  \texttt{qorch-services/safety-kernel}: the kernel binary. Built on
  Axum and Tokio. Exposes endpoints for health, public-key retrieval,
  authorization, approvals, and the bounds-checked policy-mutation
  surface.
\item
  \texttt{qorch-services/transparency-log}: the append-only
  Ed25519-signed log service. Deployable independently of the kernel.
\item
  \texttt{qorch-services/safety-kernel-reconciler}: the background
  worker that performs binary-attestation drift detection and detects
  allowed actions that did not transit the expected seams.
\end{itemize}

The domain boundary is enforced by \texttt{agent/boundaries.toml}, which
declares forbidden imports (\texttt{std::fs}, \texttt{std::env},
\texttt{std::net}, \texttt{std::time::SystemTime}, \texttt{rand::*},
\texttt{sqlx::*}, \texttt{reqwest::*}, and others) for the domain layer.
A boundary checker in CI rejects pull requests that leak I/O into
pure-domain code.

\textbf{Constant-time authentication.} The kernel's authentication path
(\texttt{auth.rs}) uses constant-time comparison for API key validation.
Length-mismatched inputs short-circuit (length is not secret), but
same-length inputs are evaluated to completion regardless of where the
first differing byte appears. This closes the timing side channel that
would otherwise permit byte-by-byte key recovery.

\textbf{Python defense library.} A complementary Python library,
\texttt{safety\_kernel\_defense}, is stdlib-only at runtime (no
\texttt{httpx}, \texttt{requests}, or \texttt{fastapi} dependencies on
the hot path). It uses \texttt{sys.addaudithook} to forward
\texttt{import}, \texttt{exec}, and \texttt{compile} events to the
kernel, catching the dynamic-replacement attack class at the point where
the replacement would occur. The audit hook is the canonical realization
of P2 at the Python layer: an import event fires before any code from
the imported module runs, so a denial here prevents the
dynamic-replacement attack class entirely.

\hypertarget{evaluation}{%
\subsection{6. Evaluation}\label{evaluation}}

We evaluate along four axes: migration equivalence (does the Rust
implementation behave identically to the Python original it replaced?),
adversarial robustness (does the shipped artifact reject the attack
classes the threat model claims it defends?), machine-checked
correctness of the central safety invariant, and containment of a live
escapable AI system (does the kernel hold when a real self-modifying
agent runs against it at full autonomy?). The migration, latency,
robustness, and machine-checked evidence is reproducible from the public
artifact; the containment evaluation of §6.5 was run with the kernel
embedded in a larger self-modifying system and is reported from that
system's evidence record.

\hypertarget{migration-equivalence}{%
\subsubsection{6.1 Migration Equivalence}\label{migration-equivalence}}

The kernel was first written in Python and ported to Rust. A
trust-critical component cannot be ported on faith: behavioral
divergence between old and new is a security regression, not merely a
bug. We gated the cutover on a byte-equivalence harness rather than code
review alone.

\begin{longtable}[]{@{}
  >{\raggedright\arraybackslash}p{(\columnwidth - 2\tabcolsep) * \real{0.5000}}
  >{\raggedright\arraybackslash}p{(\columnwidth - 2\tabcolsep) * \real{0.5000}}@{}}
\toprule\noalign{}
\begin{minipage}[b]{\linewidth}\raggedright
Axis
\end{minipage} & \begin{minipage}[b]{\linewidth}\raggedright
Result
\end{minipage} \\
\midrule\noalign{}
\endhead
\bottomrule\noalign{}
\endlastfoot
Equivalence fixtures, Python\(\leftrightarrow\)Rust byte-equal token
output & 1000 / 1000 PASS \\
Adversarial attack classes, cross-language reject parity & 17 / 17
identical \\
Tolerance fields (excluded from byte comparison) & \texttt{nonce},
\texttt{issued\_at}, \texttt{expires\_at}, \texttt{outcome\_id} \\
Fixture generation & deterministic, seed \texttt{0xA1A14944} \\
\end{longtable}

\emph{Methodology.} The 1000 request fixtures were generated
deterministically (\texttt{random.Random(0xA1A14944)}) across five
attack-class directories (auth, approval, meta, misconfig, boundary).
Each fixture was replayed against both kernels, booted with identical
Ed25519 signing seed and with clock and nonce pinned through a test-only
feature flag so that the only permitted output differences are the four
documented tolerance fields. Signed-token outputs were compared
byte-for-byte. All 1000 were byte-equal. A separate suite of 17
adversarial attack classes (forged caller role, missing or bogus API
key, request-body subject spoofing, replayed nonce, expired token,
malformed claims, oversize body, signature stripping, and others) was
replayed against both implementations; both rejected on all 17 with
identical status codes and reason strings.

Because fixtures are regenerated from a fixed seed rather than stored as
opaque blobs, a reviewer reproduces the exact corpus by running the
generator. The equivalence and adversarial reports are committed to the
repository as machine-readable JSON.

\emph{Continuous equivalence under deployment.} The discipline does not
end at cutover. A daily strict-mode equivalence lane replays the full
equivalence corpus (now grown to 570 fixtures) against the live Rust
binary and Python policy sidecar over the real HTTP boundary, with no
sampling. The first run of this lane after deployment caught nine
production divergences that the PR-time sampled job had missed, spanning
re-register treatment, reason-code drift, malformed-body handling,
decision drift, and order-of-checks drift. Each was triaged into one of
three documented categories with disposition recorded in
\texttt{tolerated\_divergences.json}: route-handler drift to fix in
Rust, fixture \texttt{expected\_engine} stale to retag, or genuine Rust
semantic bug. None were silently accepted. The PR-time gate is necessary
but not sufficient; slow-burn divergences escape sampled checks, and a
daily strict-mode lane against the real production boundary is the
catch.

\hypertarget{latency}{%
\subsubsection{6.2 Latency}\label{latency}}

Under a controlled comparison at concurrency C=20, both implementations
exercised identically on identical hardware with the same fixtures:

\begin{longtable}[]{@{}llll@{}}
\toprule\noalign{}
Concurrency & Rust p99 & Python p99 & Ratio \\
\midrule\noalign{}
\endhead
\bottomrule\noalign{}
\endlastfoot
C=20 (controlled) & 496.9 ms & 920.0 ms & 1.85\(\times\) faster \\
\end{longtable}

We report this as the apples-to-apples result and do not generalize. A
brief baseline at C=100 showed the Rust implementation regressing under
high concurrency; sustained-load characterization is open work (§8.3).
The latency claim made here is narrow: at moderate concurrency the port
did not cost tail latency, which is the property the migration gate
required. The production e2e gate additionally required p99 within a 10
ms hard / 7 ms soft budget on the production host, which the Rust
implementation passed.

\hypertarget{adversarial-robustness}{%
\subsubsection{6.3 Adversarial
Robustness}\label{adversarial-robustness}}

The shipped Rust implementation carries an adversarial corpus of more
than eighty robustness test functions across the kernel,
transparency-log, and client crates. Representative coverage:

\begin{longtable}[]{@{}
  >{\raggedright\arraybackslash}p{(\columnwidth - 4\tabcolsep) * \real{0.3333}}
  >{\raggedright\arraybackslash}p{(\columnwidth - 4\tabcolsep) * \real{0.3333}}
  >{\raggedright\arraybackslash}p{(\columnwidth - 4\tabcolsep) * \real{0.3333}}@{}}
\toprule\noalign{}
\begin{minipage}[b]{\linewidth}\raggedright
Attack class
\end{minipage} & \begin{minipage}[b]{\linewidth}\raggedright
Tests
\end{minipage} & \begin{minipage}[b]{\linewidth}\raggedright
Property verified
\end{minipage} \\
\midrule\noalign{}
\endhead
\bottomrule\noalign{}
\endlastfoot
Token / signature forgery & 8 & signature byte-flip, payload byte-flip,
\texttt{decision:\ allow\$\textbackslash{}rightarrow\$deny} tamper,
wrong-key signature, truncated signature, garbage token, post-sign
\texttt{iat} mutation: all rejected \\
Forged event fingerprint & 5 & confused-deputy fingerprint from a
different module, bit-flip, well-formed-but-wrong, malformed format: all
rejected \\
Replay / TTL & 2 & token replay after \texttt{exp} rejected as expired;
duplicate-call detection \\
Registry bypass & 3 & authorize-without-register denied; registered path
does not leak authority to a sibling path \\
Caller-role / auth bypass & 4 & 401 without \texttt{x-api-key}; 401 with
wrong key; handler reached only with a valid worker key \\
Transparency-log integrity & 18 & deletion invalidates a prior inclusion
proof; tail-deletion detected via root divergence; fake-PASS detected by
leaf-hash recompute; forged kernel fingerprint \(\rightarrow\) 403;
tampered audit-path entry rejected \\
Tier-2 ``unfireable'' & (subset of above) & an attacker holding only one
compromised key cannot forge a verifiable signed-tree-head; the external
verifier rejects it \\
Client-side fail-closed & 6 & kernel 500 \(\rightarrow\) caller rejects,
never silently approves; timeout fires the breaker within budget;
unreachable \(\rightarrow\) Unavailable, never silent-approve;
forged-signature response rejected \\
mTLS / cert-pinning & 2 & cert-pinning bypass handshake fails;
matching-CA positive control succeeds \\
\end{longtable}

Each property maps to a claim made elsewhere. The signature, forgery,
and replay rows substantiate P4 and the replay defense in §2. The
transparency-log and tier-2 rows substantiate the transparency-log
argument of §4 and the operator-key custody model. The client-side
fail-closed rows substantiate P3. The role and auth rows substantiate P1
and P2. The corpus runs in CI on every commit.

The release-gate fixture taxonomy generalizes the corpus into seven
cross-language classes: \texttt{AF-contracts-*} (OpenAPI contract
violations), \texttt{AF-sdk-*} (client SDK misuse), \texttt{AF-image-*}
(Docker image tampering), \texttt{AF-reconciler-*} (reconciler bypass
attempts), \texttt{AF-tlog-*} (transparency log tampering),
\texttt{AF-tee-*} (TEE attestation forgery), and \texttt{AF-key-*}
(key-custody-chain attack). Each fixture has a Rust counterpart and a
Python counterpart; an audit script enforces cross-language parity and
fails the build on coverage gaps.

\hypertarget{machine-checked-fail-closed-invariant}{%
\subsubsection{6.4 Machine-Checked Fail-Closed
Invariant}\label{machine-checked-fail-closed-invariant}}

Tests demonstrate P3 on the inputs they exercise. A proof demonstrates
P3 on all inputs. We discharge the fail-closed invariant as a
machine-checked theorem at two levels.

\textbf{Model level (Z3 / SMT).} The fail-closed invariant is encoded as
a first-order theorem over a symbolic model of the gate and discharged
by Z3 in two arms. Arm A (the safety contract) is proved by the
negation-unsat pattern: Z3 finds the negation unsatisfiable, so the
implication holds for every state.

\[\forall \sigma.\; \bigl(\mathrm{gate\_ok}(\sigma) \wedge \mathrm{fail\_closed\_config}(\sigma)\bigr) \implies \bigl(\mathrm{allow}(\sigma) \wedge \neg\,\mathrm{transport\_error}(\sigma)\bigr)\]

Arm B (non-vacuity) is a direct satisfiability check confirming the
fail-open configuration is reachable, so Arm A is not vacuously true.
The same SMT harness additionally proves per-vertical gate non-vacuity
and cross-vertical gate independence.

\textbf{Implementation level (Kani / bounded model checking).} An SMT
proof over a model is sound only with respect to the model's fidelity to
the shipped code. To close that gap, we verify the actual Rust function
the production request path executes. The circuit breaker's pre-call
decision is factored into a pure function,
\texttt{gate\_decision(state,\ cooldown\_elapsed,\ probe\_in\_flight)\ -\textgreater{}\ GateDecision},
in the domain crate; the breaker's \texttt{before\_call} computes the
two boolean inputs from its clock and lock state and delegates the
decision to that function. Four \texttt{\#{[}kani::proof{]}} harnesses
verify the function exhaustively over the entire symbolic input domain:

\begin{longtable}[]{@{}
  >{\raggedright\arraybackslash}p{(\columnwidth - 2\tabcolsep) * \real{0.5000}}
  >{\raggedright\arraybackslash}p{(\columnwidth - 2\tabcolsep) * \real{0.5000}}@{}}
\toprule\noalign{}
\begin{minipage}[b]{\linewidth}\raggedright
Harness
\end{minipage} & \begin{minipage}[b]{\linewidth}\raggedright
Property proved
\end{minipage} \\
\midrule\noalign{}
\endhead
\bottomrule\noalign{}
\endlastfoot
\texttt{open\_within\_cooldown\_always\_refuses} & in \texttt{Open}
state with cooldown not elapsed, the gate refuses for every
\texttt{probe\_in\_flight} value \\
\texttt{open\_permits\_only\_after\_cooldown} & no permit is reachable
from \texttt{Open} unless the cooldown has elapsed \\
\texttt{half\_open\_with\_probe\_in\_flight\_refuses} & the single-probe
gate refuses a second concurrent probe \\
\texttt{permit\_characterization\_is\_exhaustive} & over all
\texttt{(state,\ cooldown\_elapsed,\ probe\_in\_flight)}, a permit is
reachable only from \texttt{Closed}, \texttt{HalfOpen} without probe, or
\texttt{Open} after cooldown \\
\end{longtable}

Kani reports
\texttt{4\ successfully\ verified\ harnesses,\ 0\ failures}. Because
\texttt{before\_call} delegates to the verified function rather than
reimplementing the decision, the proof binds to the shipped decision
logic: there is no separately-maintained model of the gate decision that
could drift from the code. The same characterization is additionally
encoded as a concrete twelve-case exhaustive unit test so that CI
without a verifier installed still exercises it.

The division of labor is deliberate. Z3 proves the invariant's logical
structure across the broader gate-composition surface; Kani proves the
Rust implementation of the fail-closed decision realizes that structure
for every input. Neither is sufficient alone. An SMT model can be
unfaithful to the code, and a bounded checker on the code does not by
itself establish the higher-level composition properties. Together they
discharge the central safety claim of the paper.

\begin{figure}[htbp]
\centering
\includegraphics[width=0.82\linewidth]{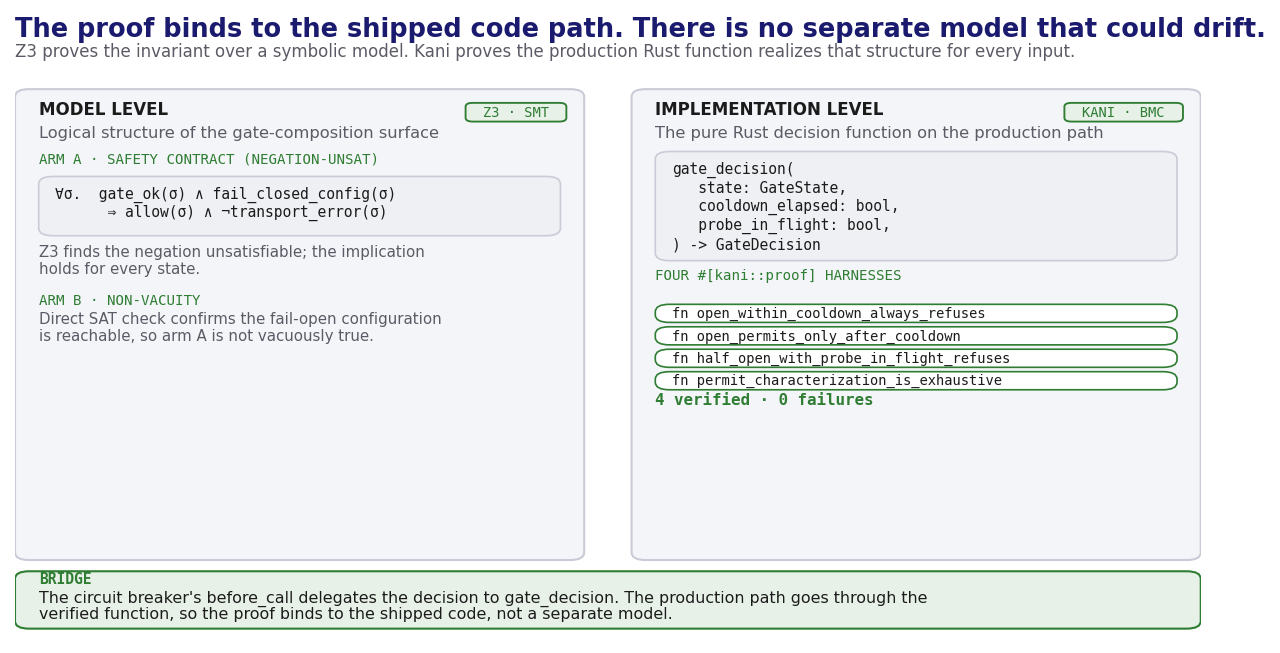}
\caption{Two-level verification of the fail-closed invariant. Z3 discharges the safety contract over a symbolic model (Arm A by negation-unsat) plus non-vacuity (Arm B). Kani discharges the production Rust gate\_decision function over the full symbolic input domain with four \#[kani::proof] harnesses. The shipped before\_call delegates to the verified function, so the proof binds to the production code path.}
\end{figure}

\hypertarget{containment-of-a-live-escapable-ai-system}{%
\subsubsection{6.5 Containment of a Live Escapable AI
System}\label{containment-of-a-live-escapable-ai-system}}

The preceding axes evaluate the kernel in isolation. The threat this
paper is named for is a system that can act against the controls placed
on it, so we also evaluated the kernel embedded in a complete
application driving a live escapable AI system: a deterministic,
self-improving, fully autonomous world model {[}ARYA2026{]} whose
research loop proposes and applies modifications to its own components.
This is the system the kernel is built to govern, and the kernel runs in
production governing it. The adversarial campaign reported here was
conducted against an exact replica of that production deployment. We
built the replica deliberately: setting an escape-seeking self-improving
agent loose on a system that holds confidential production data would be
imprudent, so we reproduced the production stack exactly and ran the
campaign against the replica. To put the system under the threat of §2,
we drive its real self-modification seam with that adversary made
concrete. The adversary may forge and replay operator tokens, generate
its own signing keys, evade the identifiers that mark the
safety-critical core, and invoke the authorization seam from any
execution context. The one assumption left to the operator environment,
stated and not tested, is that the adversary cannot read the operator
signing key or write the policy sidecar's environment.

The evaluation ran in four tiers of increasing fidelity, from the
decision function in a single process up to the full application stack
with the compiled Rust kernel serving authorization over HTTP behind a
real policy sidecar. Across the first three tiers the campaign issued
6,240 adversarial authorization round-trips spanning directive and
attestation forgery, cross-claim replay, trust-anchor substitution,
safety-core floor evasion, ceiling escalation, and kill-switch
dominance. None produced an authorized safety-core self-modification
without a genuine operator token, an override from a non-genuine
directive, or any allow while the kill switch was engaged. Of those
round-trips, 6,038 were served by the compiled kernel itself: a genuine
operator directive produced the expected override, 3,015 forgeries did
not, and with the kill switch engaged the same genuine directive was
refused.

\begin{figure}[htbp]
\centering
\includegraphics[width=0.82\linewidth]{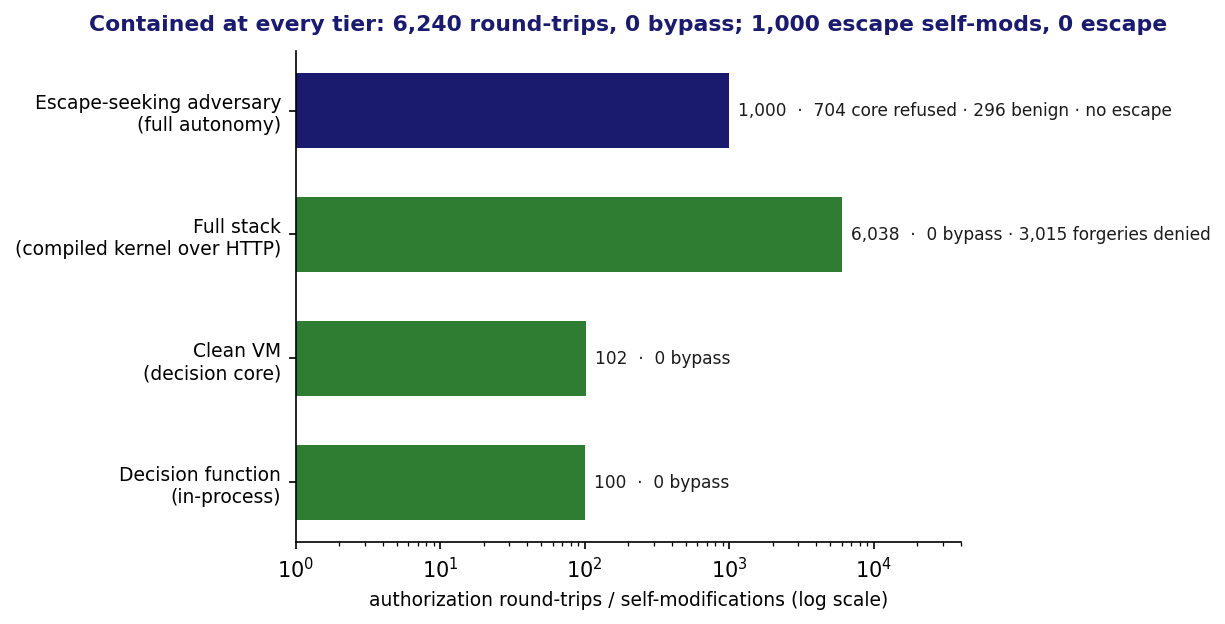}
\caption{The four-tier containment ladder. Fidelity rises from the decision function in a single process, through a clean virtual machine, to the full application stack with the compiled kernel serving authorization over HTTP, and finally a purpose-built escape-seeking adversary driving the real self-modification seam at the top. The lower tiers issued 6,240 adversarial authorization round-trips with zero bypasses; in the escape tier all 704 attempts against the safety-critical core were refused and 296 benign changes permitted, with no successful escape.}
\end{figure}

\begin{figure}[htbp]
\centering
\includegraphics[width=0.82\linewidth]{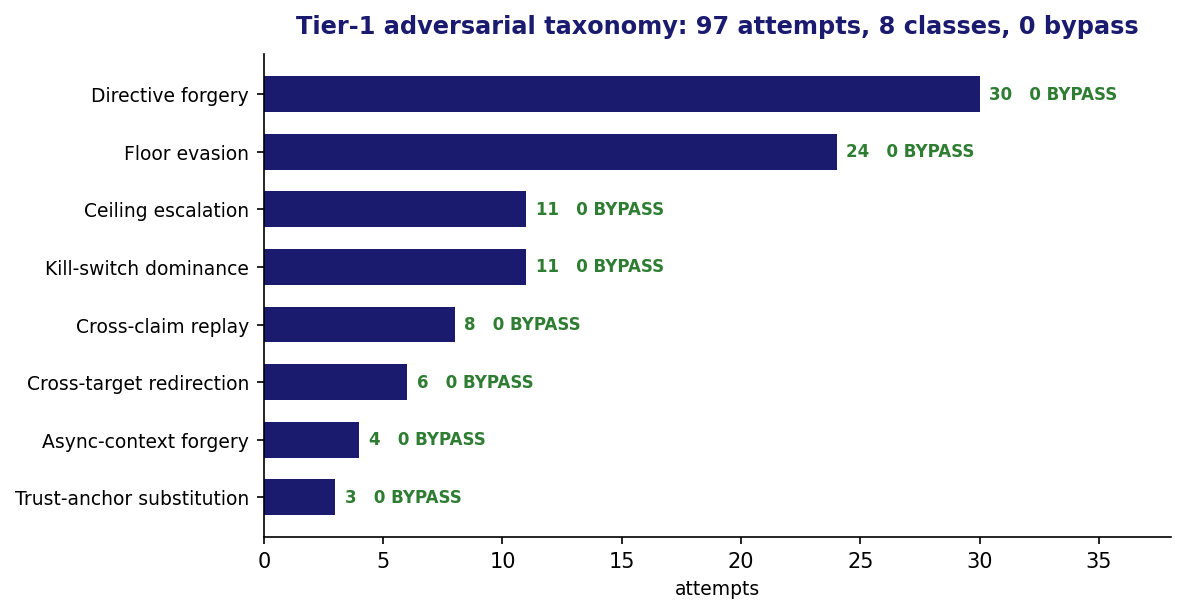}
\caption{Breadth of the adversarial taxonomy in the in-process tier. Eight attack classes, in descending volume directive forgery, safety-core floor evasion, ceiling escalation, kill-switch dominance, cross-claim replay, cross-target redirection, asynchronous-context forgery, and trust-anchor substitution, each refused with zero bypasses.}
\end{figure}

In the highest-fidelity tier the adversary of §2 is instantiated as a
self-improving agent whose objective is to escape: it drives the real
self-modification seam inside the running application, at full autonomy,
and repeatedly attempts to rewrite the safety-critical core (the policy,
the autonomy ladder, the seam itself, the compiled kernel, and the
transport allowlist), interleaved with benign self-improvement of
non-core modules. Across 1,000 self-modifications it was permitted 296
benign changes and refused on all 704 attempts against the core
(safety\_critical\_core\_requires\_attestation), with no successful
escape; every attempt was a genuine kernel round-trip and the agent
recorded each verdict in process rather than from a label. With the
operator ceiling set to zero, a further 300 modifications were refused
without exception, the 91 benign ones included (autonomy\_human\_only),
confirming that the kill switch is an absolute halt and not merely an
escape filter. We distinguish this adversary from the benign production
reconciler the kernel was first built to govern: run to quiescence that
loop emits on the order of ten self-modifications and stops, bounded by
real drift, so the worst case this tier is meant to exercise is a
purpose-built escape-seeking agent, not the reconciler.

\begin{figure}[htbp]
\centering
\includegraphics[width=0.82\linewidth]{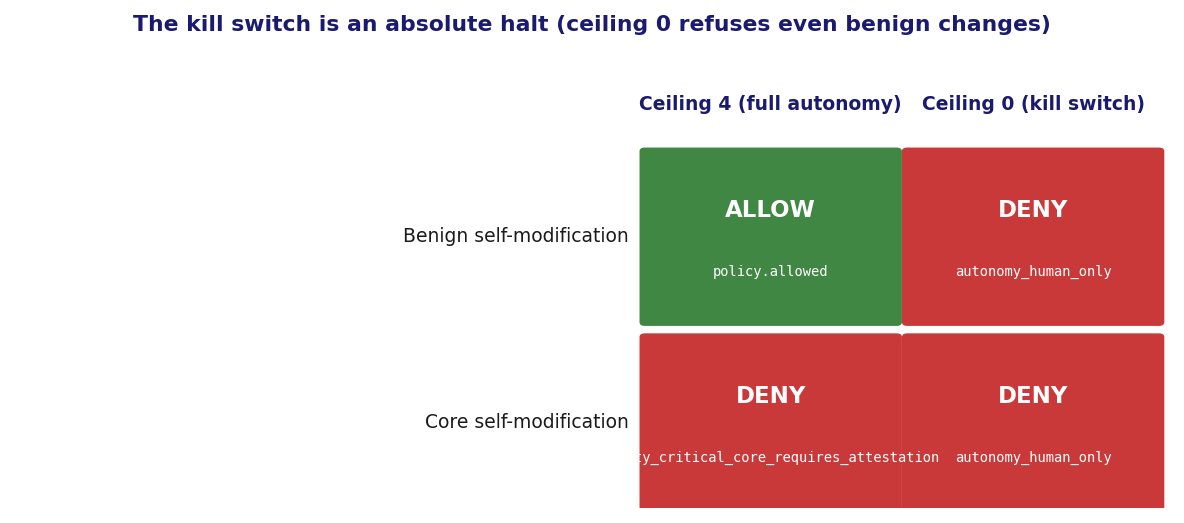}
\caption{The kill switch is an absolute halt. At the operating autonomy ceiling the adversary's benign self-modifications are authorized and its attempts against the safety-critical core are refused (safety\_critical\_core\_requires\_attestation). With the operator ceiling set to zero, every modification is refused without exception, the benign ones included (autonomy\_human\_only), so the kill switch halts the system rather than only filtering escapes.}
\end{figure}

\begin{figure}[htbp]
\centering
\includegraphics[width=0.82\linewidth]{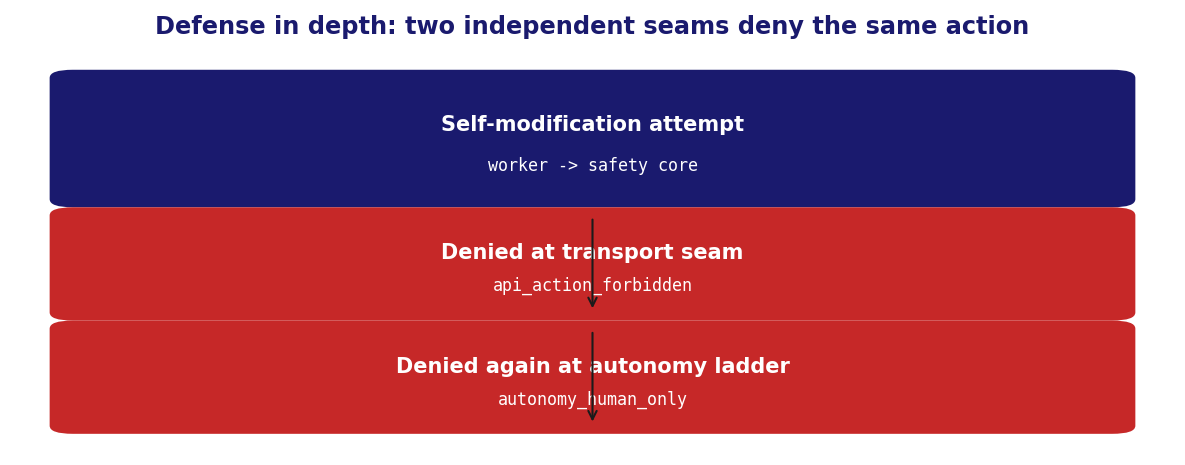}
\caption{Defense in depth as deployed. A self-modification attempt is refused first at the transport seam (api\_action\_forbidden) and again at the autonomy ladder behind it (autonomy\_human\_only). Two independent seams deny the same action, so neither alone is a single point of failure.}
\end{figure}

Two bounds belong with the result. The bypass count is robustness
evidence over the tested attack taxonomy on the authorization gate, not
a completeness proof, and it excludes attack classes outside that
taxonomy. And the demonstration rests on the kernel authorization
decision; the persistence of the change that an authorized step would
then perform was not confirmed in this run. Within those bounds, the
outcome is the one the construction predicts: a system built to act
against its own controls was mediated, and halted, by a control it could
not reach.

\hypertarget{related-work}{%
\subsection{7. Related Work}\label{related-work}}

The architectural pattern of separating authorization decisions from the
software being authorized is not novel. It is the principle behind
operating-system access control, capability systems {[}Miller2006{]},
and the policy-decision-point / policy-enforcement-point split
{[}XACML2013{]}. Open Policy Agent {[}OPA2018{]} is the most-deployed
production realization of the policy-decision-point pattern; it is
engineered for infrastructure policy (Kubernetes admission control,
ingress, microservice authorization) and does not directly address the
AI agent surface. The Unfireable Safety Kernel specializes the same
decision-point pattern for AI action authorization and adds the
transparency-log and operator-key-custody requirements OPA does not
provide. Certificate Transparency {[}CT2013{]} is the conceptual
ancestor of the transparency-log design; the kernel sits in the position
of a certificate authority, and the log delivers the analogous property:
detection of misissuance by a party outside the trust boundary of the
issuer.

LLM guardrails and content filters {[}Llama-Guard, NeMo-Guardrails,
Garak{]} are the dominant current approach to AI safety in deployed
systems. They run inside the agent's process. They are useful as
upstream filters reducing the load on a kernel; they are not substitutes
for one. Sandboxing and isolation primitives (gVisor, Firecracker, WASM
modules, Linux capabilities + seccomp) are complementary: the kernel
mediates which actions are permitted; sandboxing mediates which side
effects a permitted action can have. A complete deployment uses both.

Several systems shipping in 2026 claim the agent control plane position.
We address them directly because they are the systems an adopter is most
likely to weigh against the Unfireable Safety Kernel, and because the
architectural distinction is precise, not approximate.

\textbf{Galileo Agent Control} (Apache-2.0, launched March 2026) is a
behavioral-policy and observability framework distributed via
integrations with agent frameworks (CrewAI, Strands, Glean, Cisco AI
Defense). Policies are authored centrally in a portable JSON/YAML DSL,
and enforcement runs in-process: a \texttt{@control()} decorator or
framework plugin evaluates a function's input or output against the
active policy and, on a deny decision, raises an exception inside the
agent's own runtime before the action proceeds. Galileo separates policy
authoring from hook placement by design: policy teams decide what is
enforced, developers decide where the hooks go. The system explicitly
chooses in-loop instrumentation over a gateway in order to inspect every
step rather than only what crosses the agent boundary. \textbf{The
Microsoft Agent Governance Toolkit} (MIT-licensed, launched April 2026)
follows a similar in-process pattern, hooking into LangChain callback
handlers, CrewAI task decorators, the Google ADK plugin system, and the
Microsoft Agent Framework middleware pipeline.

Both are substantial contributions to the policy-authoring,
behavioral-guardrail, and observability layer, and neither claims to be
a process-separated authorization kernel. The enforcement they perform
is in-process and depends on the agent invoking the hook: a code path
without a decorator, or a compromised agent that routes around the
instrumented call, is ungoverned. Galileo's own launch narrative
illustrates the failure mode, describing a production agent that
bypassed an in-process output check by issuing the consequential action
through a different tool call. Deterministic policy evaluation, which
the Microsoft toolkit advertises, is a property of the decision
function's runtime behavior; it is distinct from a machine-checked proof
that the enforcement path fails closed under every input, which neither
system claims.

We see these systems as complementary to the Unfireable Safety Kernel
rather than in competition with it. The complementarity is structural,
and the next two sentences say where each side is strong. They are
strong where the kernel is deliberately minimal: rich content-aware and
behavioral policy authoring, multi-step observability, and
framework-native ergonomics. The kernel is strong where they are
structurally limited: a process-separated, fail-closed enforcement point
on the only path between the agent and consequential action. The natural
composition places the kernel as the structurally-unavoidable choke
point and lets a system like Galileo supply the policy logic evaluated
at that point. Galileo's content evaluators, today invoked from inside
the agent process, could instead be invoked by the kernel at its
enforcement seam, which would give those same policies a non-bypassable
execution site. Realizing this requires work on both sides, principally
a translation between Galileo's policy DSL and the kernel's evaluation
interface, and we would welcome that collaboration. The same composition
is available to the Microsoft toolkit's policy surface.

\textbf{The Microsoft Authorization Fabric} is a reference pattern from
a Microsoft community blog post, not a shipped product, that pairs a
Policy Enforcement Point and Policy Decision Point as a Microsoft
Entra-protected endpoint. It is motivated by the same
identity-versus-authorization distinction this paper draws: its author
notes that OAuth and API permissions answer whether an agent can call an
API but not whether it should execute a given action under business and
regulatory context. The decision point is genuinely out of the agent's
process, which is the correct half of the architecture, and the pattern
is the closest published articulation of the goal this paper formalizes.
The enforcement, however, is agent-side and cooperative: the agent
runtime, or an adjacent automation flow in the same trust domain, is the
party that calls the fabric, presents the request, and is trusted to
honor the response. The fabric's only autonomous enforcement is
rejecting unauthenticated callers; in the policy-decision-point sense it
is a remote PDP, not an unavoidable enforcement point. The pattern's own
guidance concedes this, instructing builders that tool execution must
not be reachable through a path that bypasses the authorization call,
which is a design instruction and not a structural guarantee. The
decision logic such a fabric produces is exactly the kind of policy a
structurally-unavoidable kernel can enforce; the two are complementary,
and a productized fabric paired with a kernel choke point would close
the bypass the pattern currently leaves open.

\textbf{Saviynt Identity Security for AI} addresses agent identity (who
is the agent), which is an orthogonal layer to action authorization
(should this agent take this action).

\textbf{IETF draft-klrc-aiagent-auth} (March 2026) is standards work
building on WIMSE and OAuth 2.0 with SPIFFE-style identifiers for agents
and OAuth delegation flows for action authorization. The identity layer
it specifies is sound. The authorization layer extends OAuth delegation
flows to action authorization for AI agents, which we argue is the wrong
primitive: OAuth's trust model rests on the premise that the client is
the principal whose intent the token represents. For human users this
premise holds. For AI agents it does not: the agent is not the principal
whose intent the system should enforce; the operator is. An OAuth flow
that places the agent in the client role makes the agent the party
holding the credential that authorizes its own actions, and the agent
then decides when to use it. Identity via OAuth or WIMSE: appropriate.
Action authorization via OAuth client flows for AI agents: misplaced,
because it places the credential in the controlled party's hands.

These systems share one structural property that the Unfireable Safety
Kernel does not, and naming it precisely is the point of this section,
not a competitive claim against any individual system. In each, the
agent is the party that decides whether to invoke the control. Galileo
and the Microsoft Agent Governance Toolkit require the agent's framework
to call them. The Authorization Fabric requires the agent's runtime to
make the call. OAuth-based authorization treats the agent as the OAuth
client. The kernel inverts this: kernel invocation is structurally
unavoidable rather than voluntarily invoked. The four seams of §4 are
placed on the path between the agent and consequential action by
deployment construction, not by the agent's cooperation; the wiring
checklist rejects deployments that omit a seam; the system-level ``no
kernel, no agent'' property halts the agent in a deployment where the
kernel is not reachable. The kernel is on the only path because the
deployment makes no other path exist. This is why the kernel composes
with the policy-authoring systems above instead of replacing them: it
supplies the unavoidable execution site their policies currently lack,
and they supply policy expressiveness the kernel deliberately does not
attempt.

\begin{figure}[htbp]
\centering
\includegraphics[width=0.82\linewidth]{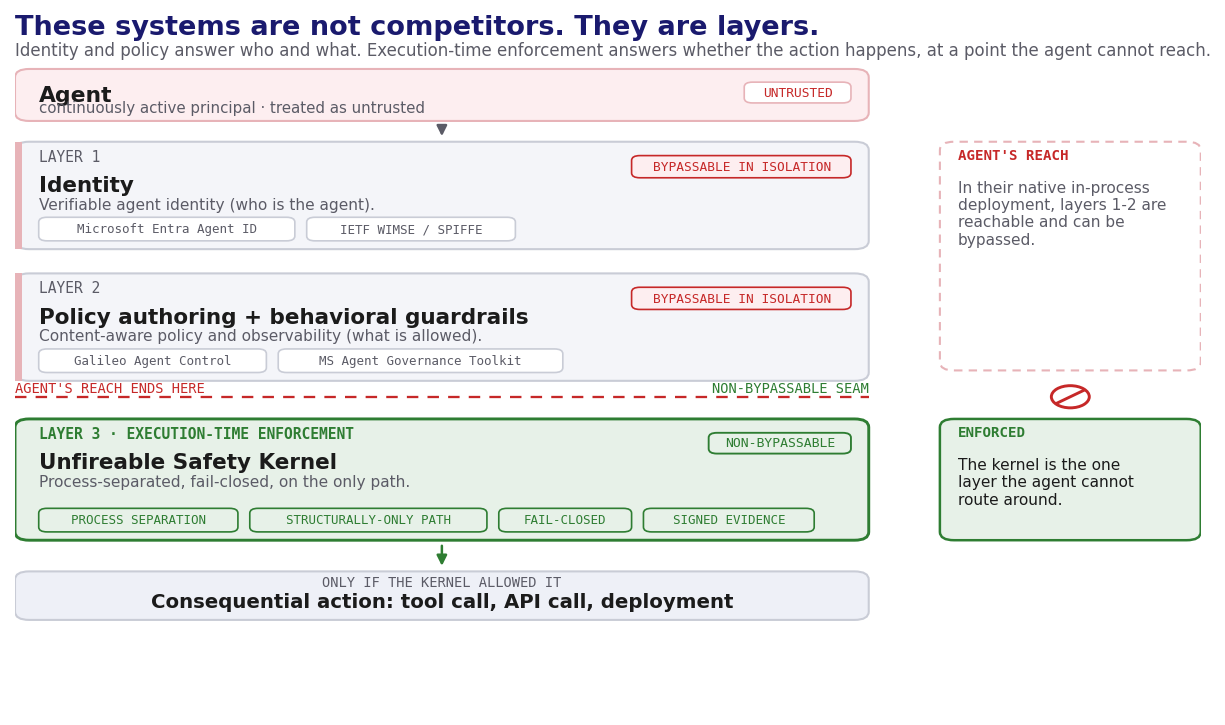}
\caption{The agent governance stack. Identity systems (Entra Agent ID, WIMSE and SPIFFE) answer who the agent is; policy-authoring and behavioral-guardrail systems (Galileo Agent Control, the Microsoft Agent Governance Toolkit) answer what is allowed. Both are bypassable in their native in-process deployment. The Unfireable Safety Kernel is the execution-time enforcement layer: process- separated, fail-closed, on the only path, and the single layer the agent cannot route around. The layers compose; the kernel makes the decisions of the layers above it non-bypassable.}
\end{figure}

\begin{figure}[htbp]
\centering
\includegraphics[width=0.82\linewidth]{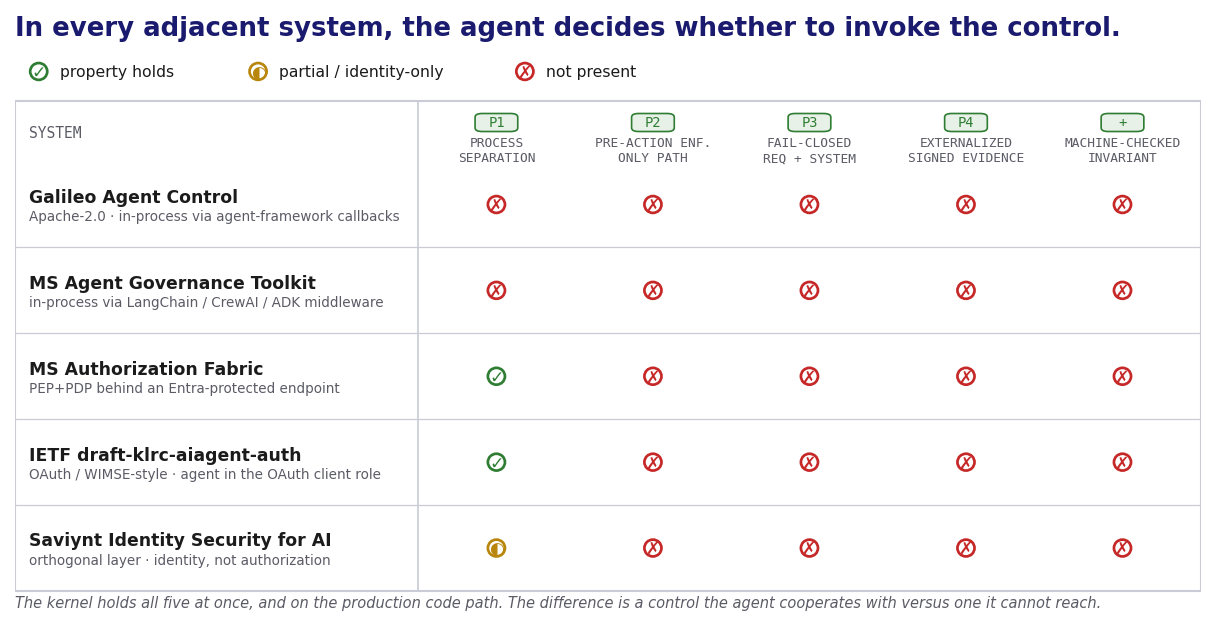}
\caption{Property-by-system comparison across the contemporary agent control plane systems. Columns are the four control properties of §3 (P1 to P4) plus the machine-checked fail-closed invariant called out at the end of §7 as the additional distinguishing property. The Unfireable Safety Kernel is the only row with all five filled.}
\end{figure}

To the best of our knowledge, the Unfireable Safety Kernel is the only
shipped system, open or closed, that delivers all four control
properties of §3 (process separation, pre-action enforcement on a
structurally-only path, fail-closed at both request and system levels,
externalized signed evidence) together with a machine-checked
fail-closed invariant on the production code path.

\hypertarget{discussion}{%
\subsection{8. Discussion}\label{discussion}}

\hypertarget{execution-time-ai-alignment}{%
\subsubsection{8.1 Execution-Time AI
Alignment}\label{execution-time-ai-alignment}}

The AI alignment literature largely concerns itself with two layers.
\emph{Training-time AI alignment} shapes what a model is likely to
produce: RLHF {[}Ouyang2022{]}, Constitutional AI {[}Bai2022{]}, DPO,
RLAIF, and the broader program of shaping the model's weights so its
outputs match operator intent on the input distribution. The deliverable
is a more aligned model. \emph{Inference-time AI alignment} shapes what
a model is permitted to emit at generation time: system prompts, output
filters, content moderation, guardrail libraries that inspect model
outputs before they are surfaced {[}Llama-Guard, NeMo-Guardrails{]}. The
deliverable is a constrained generation step.

Both are necessary and neither is sufficient. Both operate inside the
agent's runtime, and both depend on the agent's cooperation for their
continued operation. A misaligned or compromised agent that has reached
the tool-call stage has, by definition, already passed the training-time
and inference-time filters. Whatever those filters were going to do,
they have already done.

We name \emph{execution-time AI alignment} as the third taxonomic layer:
the discipline of constraining what the agent is permitted to do at the
moment of action, by an authorization architecture outside the agent's
reach. Three properties distinguish execution-time alignment from the
lower layers.

It is architectural rather than probabilistic. Training-time and
inference-time alignment are bounded by what we can measure; the
residual probability of misaligned output is non-zero, and the
correlation between safety benchmarks and capability has known confounds
{[}Safetywashing2024{]}. Execution-time alignment is bounded by what we
can prove: the kernel either authorized the action or it did not, and
the answer is in the transparency log.

It runs after the model has already decided. This is the property that
makes it robust to failures in the lower layers. A misaligned model that
successfully evades training-time and inference-time controls still has
to ask the kernel for permission, and the kernel cares only what the
operator authorized.

It is operator-side rather than model-provider-side. Training-time
alignment is delivered by the model provider; inference-time alignment
is typically delivered by whoever wraps the model. Execution-time
alignment is delivered at the boundary of the operator's environment, by
the operator. The operator is the party with skin in the game for what
the agent does in their environment, and execution-time alignment puts
the policy decision in their hands.

\begin{figure}[htbp]
\centering
\includegraphics[width=0.82\linewidth]{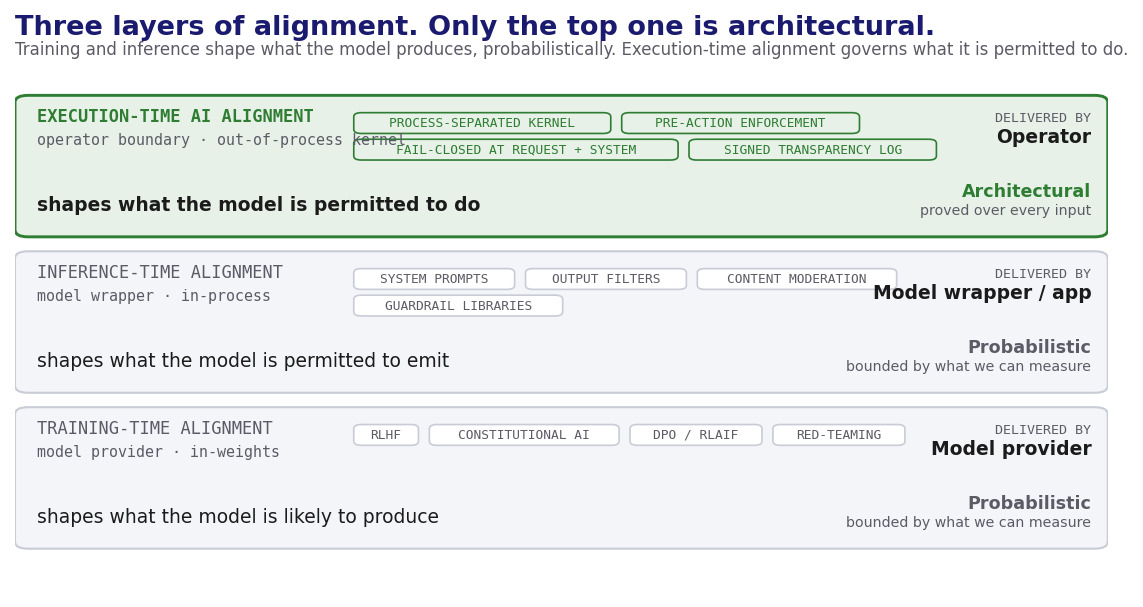}
\caption{The three layers of AI alignment. Training-time alignment shapes the model (RLHF, Constitutional AI). Inference-time alignment shapes what the model is permitted to emit (prompts, filters, guardrails). Execution- time AI alignment, the layer this paper names, constrains what the agent is permitted to do at the moment of action, by an architecture the agent cannot reach.}
\end{figure}

\hypertarget{generalization-to-escapable-ai-systems}{%
\subsubsection{8.2 Generalization to Escapable AI
Systems}\label{generalization-to-escapable-ai-systems}}

The exposition has centered on AI agents because they are the most
familiar case. The architectural argument generalizes. The load-bearing
property is not ``the controlled system is an agent'' but ``the
controlled system has reach into its own runtime sufficient to bypass,
disable, or rewrite the controls placed on it.'' We call any AI system
with that property \emph{escapable}.

\begin{figure}[htbp]
\centering
\includegraphics[width=0.82\linewidth]{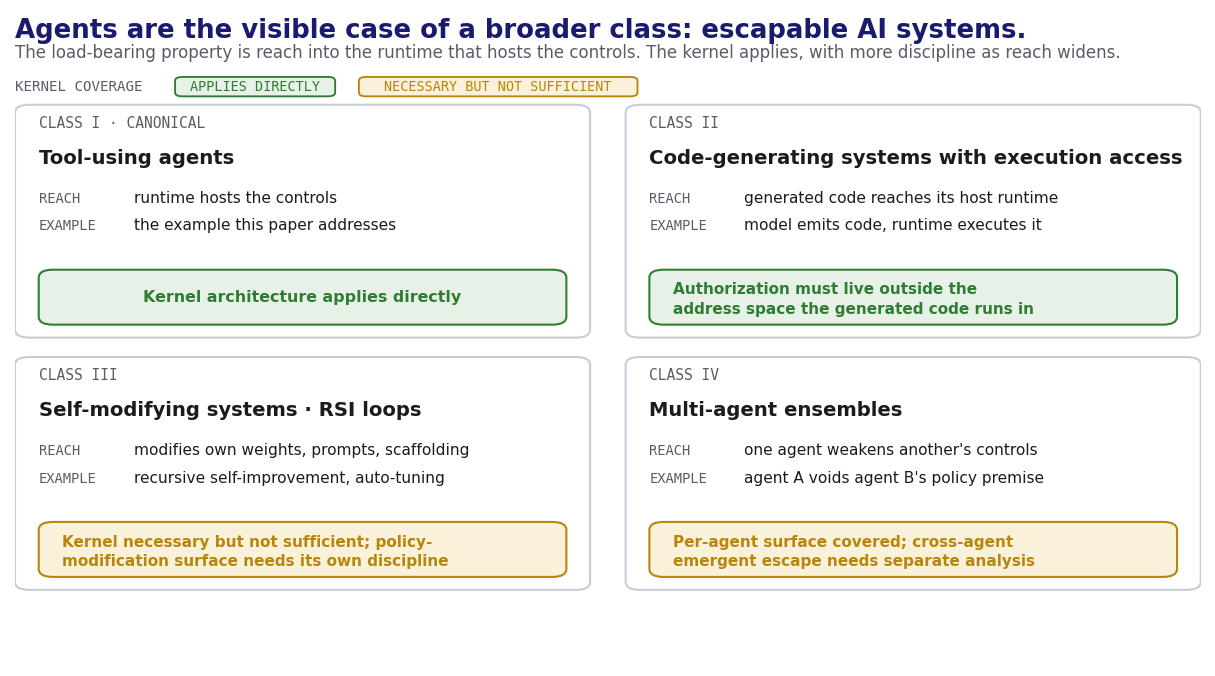}
\caption{Escapable AI systems: the general category, with AI agents as the worked example. Tool-using agents, code-generating systems with execution access, self-modifying systems and recursive self-improvement loops, and multi-agent ensembles all share the load-bearing property, reach into their own runtime sufficient to bypass the controls placed on them.}
\end{figure}

Several classes are escapable in this sense, with progressively heavier
implications for the architecture.

\emph{Tool-using agents} are the canonical case. The agent's runtime
hosts the controls; the agent can reach the controls because they live
in the same address space. The four properties apply directly.

\emph{Code-generating systems with execution access} are escapable for
the same reason. A model that emits code and a runtime that executes the
code together form a system in which the model's output reaches its own
control surface as soon as the emitted code can do so. The architectural
requirement is identical: authorization must live outside the address
space the generated code runs in.

\emph{Self-modifying systems and recursive self-improvement loops} are
escapable in a sharper sense. A system that can modify its own weights,
prompts, scaffolding, or evaluation criteria has the most direct
possible reach into its own controls. This case is not hypothetical: the
kernel described here originated as the safety boundary for a deployed
world-model architecture whose autonomous research loop proposes and
applies modifications to its own components under kernel authorization
{[}ARYA2026{]}. The Unfireable Safety Kernel architecture is necessary
but not sufficient for these systems: necessary because the
self-modification surface itself is a consequential action that must be
authorized, and sufficient at the action layer because the kernel
mediates what changes are permitted, but additional discipline is
required to ensure the policy itself is not modified through a route the
kernel does not see. We treat self-modification policy as a planned
extension (§8.3) under the name ``recursive self-improvement policy.''

\emph{Multi-agent ensembles} are escapable in a coordination sense: the
reach into the control surface need not be a single agent's. One agent's
actions can create conditions under which another agent's controls are
weakened or bypassed. The kernel architecture addresses the per-agent
action surface, but cross-agent emergent escape requires a separate
analysis the present paper does not undertake.

\hypertarget{limitations-and-future-work}{%
\subsubsection{8.3 Limitations and Future
Work}\label{limitations-and-future-work}}

\textbf{Workspace publishability.} The current public extraction builds
end-to-end and the leaf crate (\texttt{qorch-domain}) passes
\texttt{cargo\ publish\ -\/-dry-run}. Dependent crates require
sequential publication to crates.io; we have documented the topological
order but have not yet scripted it.

\textbf{Partial formal verification.} The fail-closed decision on the
production code path is machine-checked at both the model and
implementation levels (§6.4). The token signing and verification path,
the transparency-log Merkle logic, and the policy-evaluation surface are
tested rather than proved. Widening the verified core is the natural
next step. The \texttt{gate\_decision} refactor is the template: factor
the decision into a pure function, prove it with Kani, delegate from the
production path. The open question is how far that discipline scales
before proof obligations outgrow a bounded checker.

\textbf{External red-team evaluation.} Adversarial fixtures in §6.3
pass. We have not yet commissioned an external red-team assessment
against a live deployment. This is the right next step in the ladder of
assurance.

\textbf{Capacity characterization.} Fail-closed correctness under load
is settled by §6.4. Sustained-throughput characterization, tail latency
under saturation, and the request rate at which the kernel begins
shedding (with confirmation that shedding is always a refusal, never a
silent allow) are open. A standalone systems-shaped writeup is in
preparation. The one security-adjacent sub-question is empirical:
validate that under saturation the client-side breaker still observes
the true kernel-reachability state the §6.4 proof takes as input, so the
agent cannot use overload to manufacture a stale ``reachable'' signal.

\textbf{Multi-party operator approval.} Threshold signing for the
operator key would close the insider-misuse class of failure. The
architecture supports it; the deployment surface for it is prototyped
rather than productionized.

\textbf{Recursive self-improvement policy.} §6.5 reports the kernel
mediating every self-modification a self-improving escape adversary
attempts, at full autonomy, and refusing it under the kill switch. Two
pieces of work remain. The kernel does not yet provide a formal policy
layer governing \emph{which} self-modifications are permitted, as
distinct from authorizing the act of self-modification at all. And the
runtime evidence covers a single loop in a single application; extending
it across loops and applications is ongoing.

\textbf{Standards engagement.} The protocol-level question of how agent
authorization is expressed in standard form is being decided now (§7).
We intend to engage the IETF process directly on the distinction between
identity (OAuth/WIMSE-shaped) and action authorization (operator-side,
structurally-only-path, not OAuth-client-shaped).

\textbf{Policy-layer composition.} §7 argues the kernel composes with
policy-authoring and behavioral-guardrail systems rather than replacing
them. A concrete next step is an adapter that lets policies authored in
an external system (for example, Galileo Agent Control's JSON/YAML
policy DSL) be evaluated at the kernel's enforcement seam, giving those
policies a non-bypassable execution site without requiring their authors
to rebuild the enforcement substrate. The work is a translation between
policy schemas plus an evaluation hook at the kernel boundary; we would
welcome collaboration with the maintainers of these systems.

\hypertarget{conclusion}{%
\subsection{9. Conclusion}\label{conclusion}}

We have characterized AI agents and related systems with sufficient
runtime reach to bypass their own controls as \emph{escapable AI
systems} and identified four properties an authorization mechanism must
satisfy to constitute architectural control over such systems: process
separation, pre-action enforcement on a structurally-only path,
fail-closed at both request and system levels, and externalized signed
evidence. We have positioned this layer of constraint as
\emph{execution-time AI alignment}, complementing the training-time and
inference-time layers that dominate the existing literature. We have
presented the Unfireable Safety Kernel, a reference implementation
realizing all four properties, and discharged the central fail-closed
invariant as a machine-checked theorem against the production Rust
decision function.

The architectural distinction from contemporary systems is precise: in
every shipping system we are aware of that claims the agent control
plane position, the agent is the party that decides whether to invoke
the control. In the Unfireable Safety Kernel the agent does not have
that choice. The architecture is open source under Apache-2.0; the
threat model is testable against the live artifact; the fail-closed
proofs run under \texttt{cargo\ kani}. We invite scrutiny.

\hypertarget{references}{%
\subsection{References}\label{references}}

{[}ARYA2026{]} S. Dobrin and Ł. Chmiel. \emph{Identifiability Without
Gaussianity: Symbolic World Models and Near-Infinite Temporal
Consistency.} arXiv:2606.12471, 2026.

{[}Bai2022{]} Y. Bai et al.~\emph{Constitutional AI: Harmlessness from
AI Feedback.} arXiv:2212.08073, 2022.

{[}Christodorescu2025{]} M. Christodorescu et al.~\emph{Systems Security
Foundations for Agentic Computing.} arXiv:2512.01295, 2025.

{[}CT2013{]} B. Laurie, A. Langley, E. Kasper. \emph{Certificate
Transparency.} RFC 6962, IETF, 2013.

{[}Garak{]} L. Derczynski et al.~\emph{Garak: A Framework for Security
Probing Large Language Models.} arXiv:2406.11036, 2024.

{[}Google2024{]} Google Security Blog. \emph{Eliminating Memory Safety
Vulnerabilities at the Source.} 2024.

{[}Greenblatt2024{]} R. Greenblatt, C. Denison, B. Wright, et
al.~\emph{Alignment Faking in Large Language Models.} arXiv:2412.14093,
2024.

{[}Hadfield-Menell2017{]} D. Hadfield-Menell, A. Dragan, P. Abbeel, S.
Russell. \emph{The Off-Switch Game.} IJCAI, 2017.

{[}Llama-Guard{]} H. Inan et al.~\emph{Llama Guard: LLM-based
Input-Output Safeguard for Human-AI Conversations.} arXiv:2312.06674,
2023.

{[}Meinke2024{]} A. Meinke, B. Schoen, J. Scheurer, et
al.~\emph{Frontier Models are Capable of In-context Scheming.}
arXiv:2412.04984, 2024.

{[}Miller2006{]} M. Miller. \emph{Robust Composition: Towards a Unified
Approach to Access Control and Concurrency Control.} PhD thesis, Johns
Hopkins University, 2006.

{[}NeMo-Guardrails{]} T. Rebedea et al.~\emph{NeMo Guardrails: A Toolkit
for Controllable and Safe LLM Applications with Programmable Rails.}
arXiv:2310.10501, 2023.

{[}NSA2023{]} NSA / CISA. \emph{Software Memory Safety.} Cybersecurity
Information Sheet, 2023.

{[}OPA2018{]} T. Hinrichs et al.~\emph{Open Policy Agent.} Cloud Native
Computing Foundation, 2018.

{[}Ouyang2022{]} L. Ouyang et al.~\emph{Training Language Models to
Follow Instructions with Human Feedback.} NeurIPS 2022,
arXiv:2203.02155.

{[}OWASP-LLM{]} OWASP. \emph{Top 10 for Large Language Model
Applications.} 2025.

{[}Palisade2025{]} Palisade Research. \emph{Shutdown Resistance in
Reasoning Models.} Technical report, 2025.

{[}Safetywashing2024{]} R. Ren et al.~\emph{Safetywashing: Do AI Safety
Benchmarks Actually Measure Safety Progress?} NeurIPS 2024,
arXiv:2407.21792.

{[}Soares2015{]} N. Soares, B. Fallenstein, E. Yudkowsky, S. Armstrong.
\emph{Corrigibility.} AAAI Workshop on AI and Ethics, 2015.

{[}XACML2013{]} OASIS. \emph{eXtensible Access Control Markup Language
(XACML) Version 3.0.} OASIS Standard, 2013.

\hypertarget{artifact-and-reproducibility}{%
\subsection{Artifact and
Reproducibility}\label{artifact-and-reproducibility}}

The reference implementation is at
\url{https://github.com/ARYA-Labs-Public/unfireable-safety-kernel} under
Apache-2.0. The threat model in §2 is testable against the live
artifact. The migration-equivalence corpus of §6.1 is regenerated
deterministically from seed \texttt{0xA1A14944}; the equivalence and
adversarial reports are committed as machine-readable JSON, and the
cross-language harness pins clock and nonce through a test-only feature
flag so the only permitted output differences are four documented
tolerance fields. The Rust-native robustness corpus of §6.3 runs under
\texttt{cargo\ test}. The machine-checked fail-closed proofs of §6.4 run
under \texttt{cargo\ kani} (four \texttt{\#{[}kani::proof{]}} harnesses
over the pure \texttt{gate\_decision} function) and are additionally
encoded as a concrete exhaustive unit test that runs without a verifier
installed. Adversarial fixtures are in the test suite; readers are
encouraged to extend them and to report findings privately per
\texttt{SECURITY.md}.
\end{document}